\documentclass[preprint,12pt]{elsarticle}

\usepackage{amssymb}
\usepackage{amsmath}
\usepackage{lineno}
\usepackage{hyperref}
\usepackage{algorithm}
\usepackage{algpseudocode}
\usepackage{bm} 
\usepackage{float}
\usepackage{graphicx}  
\usepackage{booktabs}

\begin{document}

\begin{frontmatter}

\title{DLS: Dual-Logic Graph Neural Networks with Chaos-Fuzzy Residuals for Robust Link Prediction}

\author[1]{Mohammad Mehdi Keikha \corref{cor1}}
\cortext[cor1]{Corresponding author} 
\ead{keikha@cs.usb.ac.ir}

\author[2]{Abolfazl Nadi}

\address[1]{Department of Computer Science, University of Sistan and Baluchestan, Zahedan, Iran}
\address[2]{Department of Computer Science, School of Mathematics, Statistics and Computer Science, College of Science, University of Tehran, Tehran, Iran}

\begin{abstract}
Link prediction is a fundamental task in graph mining, critical for applications ranging from biological interaction discovery to recommender systems. While Message Passing Neural Networks (MPNNs) have achieved state-of-the-art results, they are fundamentally limited by the inadvertent suppression of high-frequency signals, as the recursive aggregation that stabilizes structural learning inherently leads to over-smoothing, rendering node representations indistinguishable in deep layers.
This limitation is particularly severe in \textit{heterophilous} graphs, where structural proximity does not imply feature similarity. In this paper, we propose \textbf{DLS (Dual-Logic SAGE)}, a novel inductive framework designed to break the over-smoothing barrier via a "controlled instability" mechanism. Unlike conventional methods relying on static structural heuristics, DLS introduces a dynamic residual pipeline. First, a \textit{Chaotic Injection Module} utilizes the Logistic Map to diversify the feature manifold, exploiting sensitivity to initial conditions to separate topologically similar but semantically distinct nodes. Second, to mitigate the aleatoric uncertainty introduced by chaotic dynamics, a \textit{Fuzzy-Logic Gating Mechanism} applies soft-thresholding to filter noise and retain only high-confidence signals. Experimental results on diverse benchmarks demonstrate that DLS effectively balances feature discrimination with robustness, significantly outperforming state-of-the-art baselines in heterophilous settings while maintaining competitive efficiency on standard homophilous graphs.
\end{abstract}


\end{frontmatter}


\section{Introduction}
\label{sec:introduction}

The comprehension of complex systems—ranging from protein-protein interaction networks in bioinformatics to social circles in digital ecosystems—relies fundamentally not merely on the constituents of the system, but on the complex structure of interactions binding them. Within the domain of graph mining, \textit{Link Prediction} has emerged as a pivotal task, aiming to uncover missing relationships or predict future interactions based on the observed state of the network \citep{wu2023graph}. The capability to accurately predict these links facilitates critical applications, including recommendation systems, drug discovery, and knowledge graph completion, making it a fundamental component of modern data science \citep{KARAMI2025}.

Historically, the evolution of link prediction techniques has traversed several distinct paradigms. The first generation relied on heuristic measures based on topological similarity, such as Common Neighbors (CN) and Adamic-Adar (AA). While computationally efficient, these methods operate under a strong assumption of triadic closure and often neglect node-specific attributes \citep{DeepLink2021}. Consequently, they fail significantly in sparse networks where local neighbors are insufficient to infer proximity. The subsequent advent of shallow embedding methods, such as DeepWalk \citep{Perozzi2014}, Node2Vec \citep{Leskovec2016} and CARE \citep{KEIKHA2018}, introduced the concept of learning latent representations via random walks. However, these approaches are inherently transductive and struggle to incorporate rich node features, limiting their applicability in dynamic, inductive scenarios where the graph structure evolves or new nodes appear.

The paradigm shifted significantly with the introduction of Graph Neural Networks (GNNs). By defining a message-passing mechanism that aggregates information from local neighborhoods, GNNs successfully unified structural topology with node attributes, establishing the current state-of-the-art \citep{yun2022graph, JWU2025}. These models operate by iteratively updating node representations, effectively projecting the graph into a low-dimensional vector space where connected nodes share similar embeddings.

Despite their prevalence, standard GNNs suffer from a fundamental theoretical limitation known as \textit{over-smoothing}. The message-passing mechanism essentially acts as a low-pass filter on the graph signal. As the network depth increases, the recursive aggregation of neighbors causes the node representations to converge towards a stationary distribution, minimizing the Dirichlet energy of the signal \citep{rusch2023survey}. This phenomenon results in the loss of high-frequency information, making nodes within the same connected component indistinguishable and severely degrading performance in deep architectures. Recent studies emphasize that bridging structural dynamics with feature learning remains the primary bottleneck in preventing this collapse \citep{GLI2025}.

This limitation is exacerbated in \textit{heterophilous} (or disassortative) graphs, where connected nodes often possess dissimilar features or labels—a common scenario in transaction networks or molecular structures \citep{Mao2025boosting}. In such environments, the homophily assumption inherent in standard GNN aggregation fails; averaging the features of dissimilar neighbors attenuates the distinctiveness of the target node, treating critical distinctive signals as noise. As noted in \citep{zheng2024graph}, standard architectures struggle to differentiate between useful heterophily and adversarial noise, leading to suboptimal generalization.

To address these challenges, we posit that the solution lies in enriching the feature space with nonlinear dynamics. We introduce \textit{Chaos Theory} into the learning process. Chaotic maps, characterized by their sensitivity to initial conditions, offer a mechanism to expand the feature manifold \citep{LIANG2025}. By subjecting node features to a chaotic mapping, we force embeddings that are artificially close (due to smoothing) to diverge in the latent space, thereby recovering their distinguishability.

However, the Incorporation of chaos introduces a trade-off: while it restores distinguishability, it may amplify aleatoric uncertainty. To mitigate this, we integrate \textit{Fuzzy Logic}. We conceptualize the link prediction task not merely as binary classification, but as determining the degree of membership in the set of existing edges. A fuzzy injection layer acts as a "Differentiable soft-thresholding mechanism," filtering out the stochastic noise introduced by chaos while preserving informative signal divergence \citep{deng2023fuzzy}.

In this paper, we propose \textbf{DLS} (Dual Logic-SAGE), a novel architecture rooted in residual learning principles that integrates these mechanisms.

\begin{itemize}
    \item \textbf{Controlled Instability Strategy:} We introduce a novel application of the Logistic Map to alleviate GNN over-smoothing by amplifying subtle feature differences, effectively increasing the resolution of the embedding space. 
    \item \textbf{Uncertainty-Aware Gating:} We propose a Fuzzy-Logic module that serves as a learnable filter, managing the trade-off between feature discriminability and signal robustness, specifically addressing the noise sensitivity issues highlighted in recent literature \citep{zheng2024graph}.
    \item \textbf{Heterophilic Robustness:} Experiments show that DLS excels in challenging heterophilous environments where traditional structural heuristics fail, validating its efficacy for inductive link prediction.
\end{itemize}
The remainder of this article is organized as follows: Section \ref{sec:related_work} contextualizes the proposed approach within the existing literature, specifically dissecting the theoretical bottlenecks of message-passing mechanisms regarding the heterophilic challenges. Section \ref{sec:methodology} formalizes the DLS framework, detailing the mathematical derivation of the chaotic injection and fuzzy gating modules within the sequential residual architecture. Section \ref{sec:experiments} presents a comprehensive empirical evaluation, including comparative analyses against state-of-the-art methods, complexity assessments, and ablation studies to validate the contribution of each module. Finally, Section \ref{sec:conclusion} synthesizes the findings and outlines potential directions for future research.

\section{Related Work}
\label{sec:related_work}

Unlike node classification which focuses on individual entities, Link Prediction (LP) requires learning the joint probability distribution of node pairs $(u, v)$. Existing literature can be categorized into three evolution stages: heuristic methods, latent feature approaches, and subgraph-based neural architectures.

\subsection{Heuristic and Latent Feature Baselines}
The earliest LP methods relied on topological heuristics such as Common Neighbors (CN), Adamic-Adar (AA), and Resource Allocation (RA) \citep{wang2023neural}. These methods operate on the premise that nodes with significant neighborhood overlap are likely to connect. While efficient, they fail in sparse graphs where overlap is minimal or non-existent.
Following this, latent feature models like Matrix Factorization and random-walk embeddings (e.g., DeepWalk \citep{Perozzi2014}, Node2Vec \citep{Leskovec2016} and CARE \citep{KEIKHA2018}) emerged. These methods project nodes into a vector space where links are predicted via geometric proximity (e.g., dot product). However, these shallow embeddings are transductive; they cannot handle new nodes in evolving networks without complete retraining, a critical flaw for real-world inductive LP applications \citep{hamilton2017inductive}.

\subsection{Deep Graph Link Prediction}
Modern approaches utilize Graph Neural Networks (GNNs) to encode structural information. Standard methods (e.g., GAE, VGAE) use GNNs as encoders and a simple inner-product as a decoder. However, simple dot-product decoding fails to capture complex structural dependencies.
To address this, recent state-of-the-art methods focus on enclosing subgraphs. BUDDY \citep{chamberlain2023buddy} and Neo-GNN \citep{yun2021neo} utilize structural sketches and adjacency overlap to explicitly model link existence. Similarly, NCN (Neural Common Neighbor) \citep{wang2023neural} integrates learnable structural features directly into the pairwise scoring function.
Despite their accuracy, these methods often suffer from high computational latency due to subgraph extraction overhead or complex structural hashing. Conversely, embedding-based GNNs like GraphSAGE \citep{hamilton2017inductive} are faster but struggle with distinguishability in deep layers, necessitating a new class of efficient, non-linear feature enhancers like DLS.

\subsection{The Pairwise Indistinguishability Limitation}
A critical issue specific to GNN-based link prediction is the \textbf{``Pairwise Indistinguishability''} problem. Recent surveys highlight that as GNN depth increases, the recursive aggregation minimizes the Dirichlet energy of the graph signal \citep{rusch2023survey}.
Mathematically, this causes node embeddings within a cluster to converge to a single point (oversmoothing). For link prediction, this is detrimental: if $h_u \approx h_v$ for all nodes in a community, the predicted probability $P(A_{uv}=1) \propto \sigma(h_u^T h_v)$ becomes uniform for all pairs. Current solutions rely on gating mechanisms or skip-connections \citep{rusch2023gradient}. DLS offers a different perspective by using chaotic dynamics to expand the feature manifold, preventing this collapse without expensive structural mining.

\subsection{Heterophily and Uncertainty in Link Prediction}
Most LP methods implicitly assume homophily (connected nodes are similar). This assumption fails in heterophilous graphs (e.g., protein interactions) where edges connect dissimilar entities. As noted in recent comprehensive surveys, standard aggregators effectively attenuate the distinctive features needed to predict such links \citep{zheng2024graph, yan2023heterophily}.
Furthermore, LP is a \textbf{Positive-Unlabeled (PU)} learning task; a missing edge represents epistemic uncertainty rather than a definite negative. While logic-based methods have been proposed, they often lack differentiability. By integrating fuzzy logic into the neural pipeline, DLS provides a gradient-friendly mechanism to gate this uncertainty, explicitly modeling the ``degree of truth'' for edge existence, similar to approaches seen in few-shot learning \citep{deng2023fuzzy}.

\
\section{Methodology}
\label{sec:methodology}

In this section, we formulate the link prediction problem and present \textbf{DLS (Dual Logic-SAGE)}, a residual learning framework designed to mitigate the over-smoothing and heterophily bottlenecks identified in Sections \ref{sec:introduction} and \ref{sec:related_work}. Unlike conventional ensembles that process features in parallel, DLS employs a \textit{Sequential Residual Refinement} strategy. This approach allows the model to progressively enrich the node embeddings: first by capturing local topology via GraphSAGE, then by injecting chaotic non-linearity, and finally by applying fuzzy logic smoothing.

\subsection{Problem Formulation}
Let $\mathcal{G} = (\mathcal{V}, \mathcal{E})$ be an undirected graph with $N = |\mathcal{V}|$ nodes and a feature matrix $X \in \mathbb{R}^{N \times F}$, where each row $x_i$ represents the attribute vector of node $i$. The task of Link Prediction is to learn a scoring function $\psi: \mathcal{V} \times \mathcal{V} \rightarrow [0, 1]$ that estimates the probability $P(A_{uv}=1 | X, \mathcal{E}_{obs})$ for a candidate pair $(u, v)$, where $\mathcal{E}_{obs}$ is the set of observed edges.

Standard GNNs estimate this via embeddings $Z$ such that $\psi(u, v) = \sigma(z_u^T z_v)$. However, deep GNNs minimize the Dirichlet energy of the graph signal, causing $z_u \approx z_v$ for local neighbors regardless of their true affinity \citep{rusch2023survey}. To resolve this, DLS redefines the embedding $Z$ not as a static projection, but as a dynamic composite of three forces: structural topology, chaotic expansion, and fuzzy logic gating.

\subsection{Phase 1: Dual-Channel Feature Projection}
\label{sec:phase1}
Before feeding data into the neural network (the message-passing phase), we construct two auxiliary views of the latent space derived from the raw node features $X$. These engineered features are crucial to capture dynamics that static topology of the network cannot reveal.

\subsubsection{Chaotic Dynamics Channel ($X_c$): Manifold Expansion}
To counter the low-pass filtering effect of GNNs, we require a mechanism to re-inject high-frequency variations. We turn to \textit{Chaos Theory}, specifically the concept of ``sensitivity to initial conditions.'' In dynamical systems, chaotic maps ensure that two points initially close in phase space diverge exponentially over time \citep{may1976simple}.

We employ the Logistic Map, a classic example of complex dynamical behavior arising from simple non-linear equations. The chaotic feature matrix $X_c$ is generated by mapping the normalized raw features $\hat{X}$ into a chaotic regime:

\begin{equation}
    \label{eq:chaos}
    X_{c} = r \cdot \hat{X} \odot (1 - \hat{X})
\end{equation}

where $\odot$ denotes the Hadamard product and $r$ is the control parameter. The $r$ parameter guarantees that the system operates in the chaotic domain (positive Lyapunov exponent  $\lambda > 0$). This transformation acts as a \textit{Manifold Expansion} operator. If two nodes $u$ and $v$ have features that are dangerously similar (leading to potential over-smoothing later), the chaotic map amplifies their minute differences. This effectively ``stretches'' the feature space, ensuring that distinct entities remain distinguishable even after aggregation, addressing the indistinguishability crisis \citep{rusch2023gradient}.

\subsubsection{Fuzzy Logic Channel ($X_f$): Uncertainty Modeling}
While chaos improves distinguishability, it inherently introduces aleatoric uncertainty (noise). To manage this, we project features into a fuzzy logic domain using a soft-membership function. We utilize the hyperbolic tangent as a continuous, differentiable approximation of a fuzzy set membership degree:

\begin{equation}
    \label{eq:fuzzy}
    X_{f} = \tanh(\alpha \cdot X)
\end{equation}

where $\alpha$ is a learnable scaling factor controlling the slope of the fuzziness. In fuzzy set theory, values close to 0 represent maximum uncertainty (neither true nor false), while values approaching $\pm 1$ represent high confidence. This channel ($X_f$) acts as a \textit{Semantic Filter}. By pushing weak signals towards 0 and saturating strong signals, it creates a robust mask that can later gate the chaotic perturbations, ensuring that only statistically significant variations influence the final prediction \citep{deng2023fuzzy}.

\subsection{Phase 2: Sequential Residual Architecture}
\begin{figure}[htbp]
    \centering
    \includegraphics[width=1.0\linewidth]{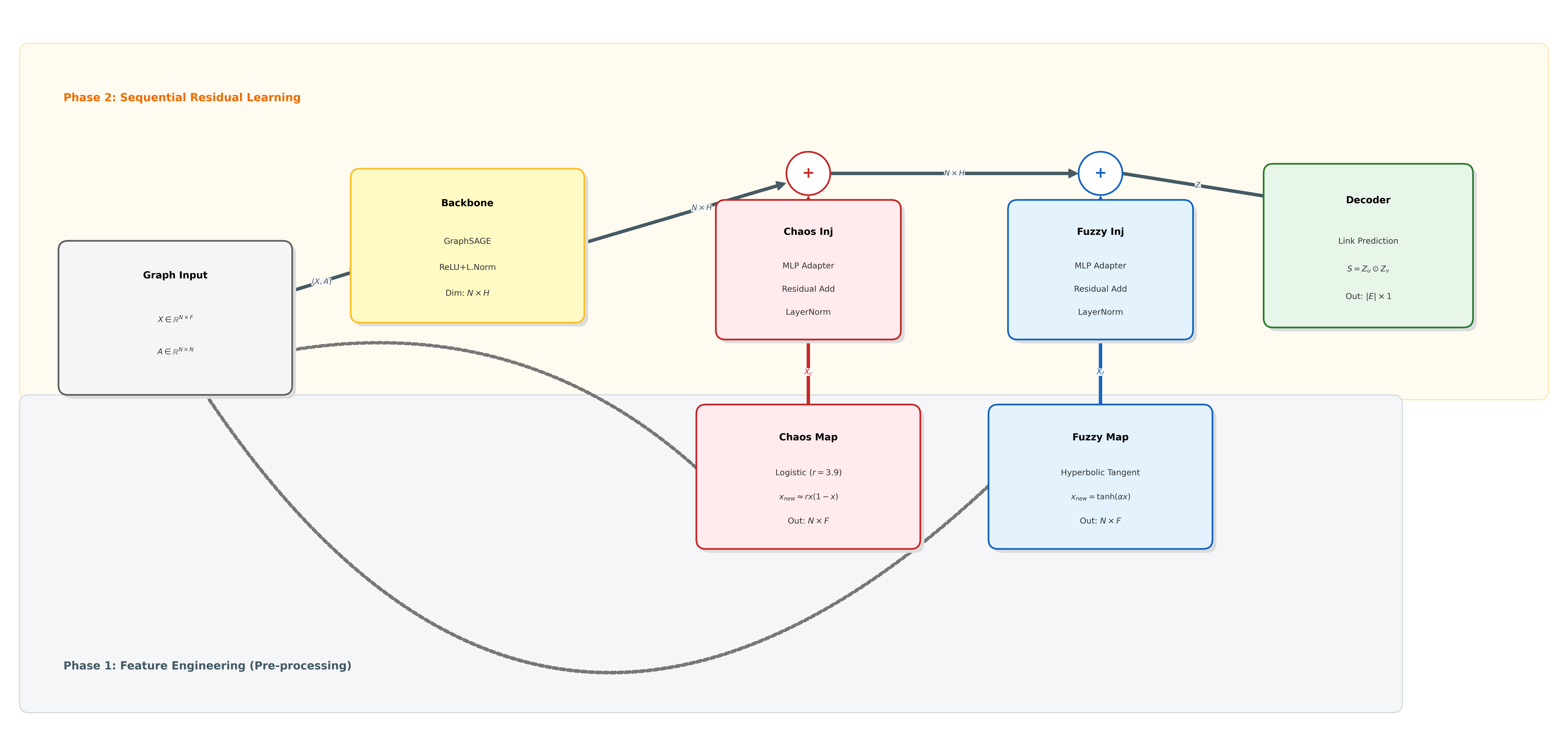}
    \caption{The proposed DLS architecture. The framework consists of a GraphSAGE backbone augmented by two residual streams: a Chaotic Injection path for feature expansion and a Fuzzy Logic path for noise gating. The final embedding $Z$ integrates structural, chaotic, and fuzzy information sequentially.}
    \label{fig:dls_arch}
\end{figure}

The core innovation of DLS is its strictly sequential processing pipeline. The architecture integrates the channels defined in Phase 1 via a cascaded residual pipeline. Crucially, the auxiliary features $X_c$ and $X_f$ are \textbf{not} concatenated to the input; instead, they are projected via learned adapters and injected into the hidden layers to refine the structural embeddings. The complete architecture is illustrated in Figure \ref{fig:dls_arch}.

The propagation process consists of the following sequential steps:

\subsubsection{Step 1: Structural Encoding (Backbone)}
The primary stream uses a standard GraphSAGE encoder to capture local homophily and structural topology. Given the input $X$ and edge list $E$, the base embedding $H_{base} \in \mathbb{R}^{N \times d}$ is computed as:
\begin{equation}
    H_{base} = \text{LayerNorm}(\text{ReLU}(\text{SAGEConv}(X, E)))
\end{equation}
At this stage, $H_{base}$ captures the general community structure but suffers from the standard over-smoothing limitations inherent in aggregation.

\subsubsection{Step 2: Chaotic Residual Injection}
To restore high-frequency details, we inject the chaotic signal $X_c$. Crucially, this is done via a Multi-Layer Perceptron (MLP) adapter ($W_c$) that acts on $X_c$ and adds the result to the existing stream as a residual term:
\begin{equation}
    H_{chaos} = \text{LayerNorm}\left( H_{base} \oplus \text{MLP}_{chaos}(X_c) \right)
\end{equation}
This operation effectively perturbs the smoothed embeddings. If two nodes $u, v$ have $h_u \approx h_v$ (indistinguishable) but distinct chaotic states, this layer forces their representations apart. The residual addition $\oplus$ ensures that the topological information ($H_{base}$) is not lost but rather sharpened.

\subsubsection{Step 3: Fuzzy Logic Refinement}
Finally, the embedding is refined using the fuzzy features $X_f$. This step mimics a logic gate, smoothing out the chaotic noise introduced in the previous step and preparing the features for classification:
\begin{equation}
    Z = \text{LayerNorm}\left( H_{chaos} \oplus \text{MLP}_{fuzzy}(X_f) \right)
\end{equation}
This final embedding $Z$ represents a structurally sound representation that is both high-resolution (via Chaos) and noise-robust (via Fuzzy Logic). The use of \textit{LayerNorm} after each residual addition is critical for maintaining numerical stability and preventing gradient explosion.

\subsection{Link Prediction Decoder}
Once the final node embeddings $Z$ are generated, the probability of a link existing between node $u$ and node $v$ is calculated. We employ the Hadamard product (element-wise multiplication) to capture the interaction between node pairs, followed by a decoding MLP which effectively captures the logical conjunction of the refined features:

\begin{equation}
    P(A_{uv}=1 | Z) = \sigma \left( \text{MLP}_{dec}(Z_u \odot Z_v) \right)
\end{equation}

This structure ensures that the prediction relies on the intersection of features refined by both chaos and fuzzy logic.

\subsection{Training and Complexity}
The model is trained end-to-end using Binary Cross Entropy (BCE) loss on the predicted edge probabilities. The complete training procedure, including the pre-processing of features and the sequential forward pass, highlights the efficiency of the DLS model.

\textbf{Complexity Analysis:} The DLS architecture is highly efficient. The Chaos and Fuzzy injections (Phase 1) are $\mathcal{O}(N)$ node-wise operations. The dominant cost remains the SAGE aggregation $\mathcal{O}(|\mathcal{E}|)$. Thus, DLS maintains linear scalability $\mathcal{O}(|\mathcal{E}| + N)$, making it significantly faster than subgraph-based methods like SEAL or NCN which scale with the number of target links.

\section{Experiments and Results}
\label{sec:experiments}
In this section, we present a comprehensive empirical evaluation of the proposed DLS framework. To validate the model, we subject the model to a rigorous testing protocol across diverse benchmark datasets, comparing its efficacy against leading state-of-the-art baselines in terms of predictive accuracy, computational scalability, and parameter sensitivity. Furthermore, we conduct an ablation study to dissect the individual contributions of the chaotic injection and fuzzy regularization mechanisms, empirically verifying their synergistic impact on the model’s overall predictive capability. To ensure consistent reproducibility across all reported metrics, all computational experiments were executed within an environment powered by an NVIDIA Tesla T4 GPU.

\subsection{Experimental Setup}
\label{subsec:setup}

\subsubsection{Datasets}
To rigorously evaluate the generalization capability and robustness of the DLS architecture, we selected four benchmark datasets that span a wide spectrum of topological characteristics, ranging from homophilic citation networks to highly heterophilic web graphs.
\begin{itemize}
    \item \textbf{Cora} \cite{sen2008collective}: A canonical homophilic citation network where edges signify academic references, typically exhibiting high feature smoothing.
    \item \textbf{Squirrel} and \textbf{Chameleon} \cite{pei2020geomgcn}: Wikipedia page networks characterized by high heterophily, where linked nodes frequently possess dissimilar feature vectors, posing a challenge for standard smoothing-based GNNs.
    \item \textbf{Actor} \cite{tang2009social}: A movie-actor co-occurrence network renowned for its low clustering coefficient and complex structural dependencies, serving as a stress test for structural learning.
\end{itemize}
This diverse selection ensures that our evaluation moves beyond simple assortative mixing scenarios, testing the model's efficacy in disassortative environments where traditional message passing often falters.

\subsubsection{Baseline Methods}
We benchmark DLS against a suite of state-of-the-art (SOTA) link prediction methodologies, categorizing them into structure-learning and efficient subgraph-sketching approaches:
\begin{itemize}
    \item \textbf{Neo-GNN} \cite{yun2022graph}: A structural augmentation method that aggregates adjacency matrix powers to capture neighborhood overlap, mitigating standard over-smoothing.
    \item \textbf{BUDDY} \cite{chamberlain2023buddy}: A scalable framework utilizing subgraph sketching and pre-computed structural hashes (RA/AA) to approximate link likelihoods without expensive subgraph extraction.
    \item \textbf{ELPH} \cite{chamberlain2022elph}: The precursor to BUDDY, employing a simplified hashing mechanism to propagate structural heuristics.
    \item \textbf{NCNC} \cite{wang2023neural}: A Neural Common Neighbor approach integrated with completion mechanisms to explicitly model and recover missing links during the training phase.
\end{itemize}

\subsection{Computational Complexity Analysis}
\label{subsec:complexity}

A pivotal advantage of the DLS architecture lies in its linear scalability, contrasting sharply with the computational bottlenecks inherent in subgraph-based or completion-based baselines. Table \ref{tab:complexity} provides a theoretical complexity analysis.

Let $|V|$ denote the number of nodes, $|E|$ the number of edges, $d_{avg}$ the average node degree, and $F$ the feature dimension.

\begin{table}[htbp]
    \centering
    \caption{Theoretical Time Complexity Comparison. DLS maintains linear complexity similar to standard GNNs, avoiding the high pre-processing costs of subgraph sketching (BUDDY) or completion (NCNC).}
    \label{tab:complexity}
    \resizebox{\linewidth}{!}{
    \begin{tabular}{lcccc}
        \toprule
        \textbf{Model} & \textbf{Pre-processing} & \textbf{Training/Inference} & \textbf{Space Complexity} & \textbf{Scalability} \\
        \midrule
        Neo-GNN \cite{yun2022graph} & $\mathcal{O}(|E|)$ & $\mathcal{O}(L|E|F + |V|F^2)$ & $\mathcal{O}(|V|F + |E|)$ & High \\
        ELPH \cite{chamberlain2022elph} & $\mathcal{O}(|E|)$ & $\mathcal{O}(|E|_{sub} \cdot B)$ & $\mathcal{O}(|V|F)$ & Medium \\
        BUDDY \cite{chamberlain2023buddy} & $\mathcal{O}(|V| \cdot d_{max}^2)$ & $\mathcal{O}(B \cdot F_{struct})$ & $\mathcal{O}(|V|F + |E|_{sketch})$ & Low/Medium \\
        NCNC \cite{wang2023neural} & - & $\mathcal{O}(|E|F + |V|F^2 + |E|^2_{complete})$ & $\mathcal{O}(|V|F + |V|^2_{overlap})$ & Low \\
        \midrule
        \textbf{DLS (Ours)} & $\mathcal{O}(|V|)$ & $\mathcal{O}(L|E|F + |V|F^2)$ & $\mathcal{O}(|V|F + |E|)$ & \textbf{Very High} \\
        \bottomrule
    \end{tabular}
    }
\end{table}

\noindent \textbf{Analysis:} 
\begin{enumerate}
    \item \textbf{Pre-processing Efficiency:} Unlike BUDDY, which necessitates computationally intensive pre-computation of subgraph sketches (often scaling quadratically with degree for structural features like RA/AA), DLS requires only a single linear pass $\mathcal{O}(|V|)$ to generate the chaotic and fuzzy auxiliary features ($X_c, X_f$).
    \item \textbf{Training Latency:} DLS introduces negligible overhead to the training loop. Unlike NCNC, which relies on neighbor completion (involving potentially dense matrix operations), DLS employs lightweight element-wise MLP injections. The complexity remains dominated by the sparse matrix multiplication in the Base Encoder ($\mathcal{O}(|E|F)$), ensuring strictly linear scalability relative to graph size.
    \item \textbf{Memory Constraints:} By circumventing the storage requirements of subgraph sketches or completed adjacency matrices, DLS effectively mitigates Out-Of-Memory (OOM) errors on large-scale graphs, a limitation frequently observed with NCNC and BUDDY in resource-constrained environments.
\end{enumerate}

\subsection{Hyperparameter Sensitivity Analysis}
\label{subsec:sensitivity}

\begin{figure*}[htbp]
    \centering
    \includegraphics[width=\textwidth]{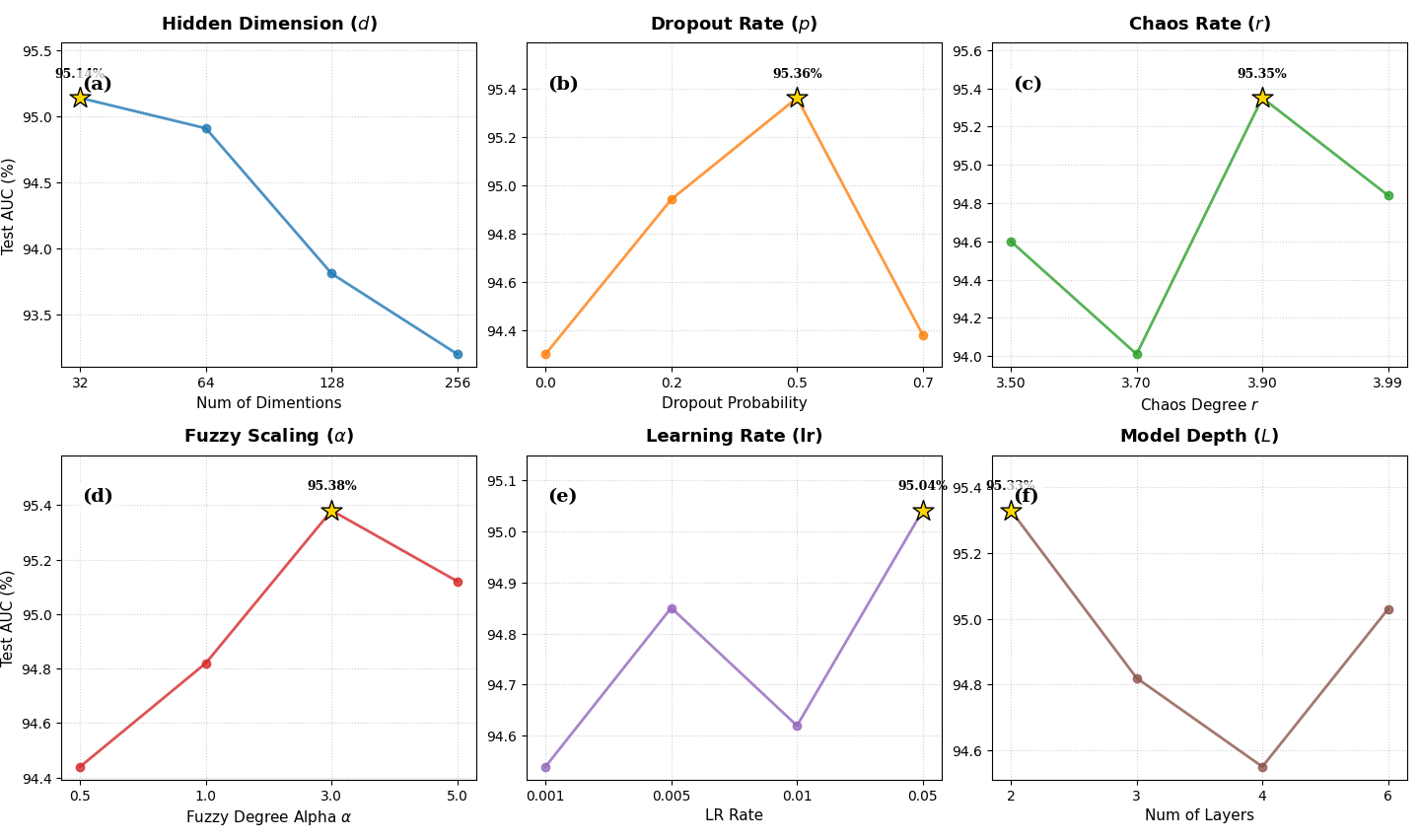}
    \caption{Hyperparameter sensitivity analysis on the Squirrel dataset. The optimal configurations are marked with a star ($\star$). (a) Hidden Dimension ($d$), (b) Dropout Rate ($p$), (c) Chaos Rate ($r$), (d) Fuzzy Scaling ($\alpha$), (e) Learning Rate ($lr$), and (f) Model Depth ($L$).}
    \label{fig:sensitivity}
\end{figure*}

To ascertain the optimal configuration for DLS, we performed a comprehensive sensitivity analysis using the \textbf{Squirrel} dataset as the representative benchmark. This choice was motivated by its high heterophily and structural complexity, which serve as a rigorous stress test for the model's dynamic components. The optimal parameters identified were subsequently fixed across all datasets to evaluate zero-shot generalization without dataset-specific tuning. We employed a standard grid search protocol. The hyperparameter landscape is visualized in Figure \ref{fig:sensitivity}.

\subsubsection{Impact of Structural Parameters ($d, L$)}
As depicted in Fig. \ref{fig:sensitivity}(a), DLS exhibits consistent parameter efficiency, achieving peak performance (95.14\%) with a compact hidden dimension of $d=32$. Notably, increasing dimensionality to 128 or 256 degrades performance. This counter-intuitive finding suggests that the chaotic injection mechanism successfully enriches the feature space with high-frequency details, allowing for high expressivity without the need for high-dimensional latent spaces, thereby mitigating the curse of dimensionality.
Regarding model depth, Fig. \ref{fig:sensitivity}(f) confirms that a shallow architecture ($L=2$) outperforms deeper variants (95.33\%). This aligns with theoretical expectations for heterophilous graphs, where long-range message passing tends to aggregate noise rather than signal, exacerbating the over-smoothing problem discussed in Section \ref{sec:introduction} \citep{rusch2023survey}.

\subsubsection{Regularization and Optimization ($p, lr$)}
Figure \ref{fig:sensitivity}(b) underscores the necessity of robust regularization. A relatively high dropout rate of $p=0.5$ yields optimal results (95.36\%), confirming that regularization is essential to prevent the model from overfitting to the stochastic "deterministic noise" introduced by the chaos module \citep{srivastava2014dropout}.
In terms of optimization dynamics, Fig. \ref{fig:sensitivity}(e) indicates that a learning rate of $lr=0.05$ strikes the best balance between convergence speed and stability (95.04\%). Lower rates result in sluggish convergence, while higher rates induce oscillation around local minima.

\subsubsection{Analysis of Dynamic Control Parameters ($r, \alpha$)}
The sensitivity profiles of the core dynamic components, shown in Figs. \ref{fig:sensitivity}(c) and (d), empirically validate our theoretical premises:

\begin{itemize}
    \item \textbf{Chaos Rate ($r$):} Figure \ref{fig:sensitivity}(c) demonstrates that performance improves as the system approaches the chaotic regime. The performance peaks at $r=3.9$ (95.35\%). In dynamical systems theory, this value corresponds to the \textit{"edge of chaos"}, where the Logistic Map exhibits maximum entropy and complexity without devolving into pure white noise \citep{may1976simple}. At $r=3.99$, the dynamics become excessively unpredictable, leading to a slight degradation in predictive accuracy.
    
    \item \textbf{Fuzzy Scaling ($\alpha$):} The fuzzy scaling parameter governs the steepness of the hyperbolic membership function. As seen in Fig. \ref{fig:sensitivity}(d), a sharper decision boundary ($\alpha=3.0$) significantly outperforms the standard scaling ($\alpha=1.0$), achieving an AUC of 95.38\%. This indicates that for link prediction, enforcing a distinct separation—effectively a soft-thresholding mechanism—between connected and disconnected fuzzy sets is crucial for filtering aleatoric uncertainty.
\end{itemize}

Based on this analysis, the configuration $\{d=32, p=0.5, r=3.9, \alpha=3.0, lr=0.05, L=2\}$ is adopted as the default setting for all subsequent comparative experiments.

\subsection{Comparative Performance Evaluation}
\label{subsec:comparative}

To rigorously evaluate the effectiveness of DLS, we benchmarked it against four advanced state-of-the-art baselines across four datasets representing both homophilous (Cora) and heterophilous (Squirrel, Chameleon, Actor) structures. The experimental setup strictly followed the protocol defined in Section \ref{subsec:setup}: we reported the average Test AUC and standard deviation over 5 independent random seeds ($seeds \in \{42, 1, 123, 0, 2026\}$). To prevent data leakage, a strict temporal split was enforced where test edges were entirely excluded from the message-passing aggregations. Table \ref{tab:main_results} summarizes the comparative results.

\begin{table*}[htbp]
\centering
\caption{Comparative performance (Test AUC in \%) on homophilous (Cora) and heterophilous (Squirrel, Chameleon, Actor) datasets. Results are reported as Mean $\pm$ Std over 5 seeds. The best performance is \textbf{bolded}, and the second best is \underline{underlined}. \textit{OOM} indicates Out-Of-Memory error on a standard 16GB GPU.}
\label{tab:main_results}
\resizebox{\textwidth}{!}{%
\begin{tabular}{lcccc}
\toprule
\textbf{Model} & \textbf{Cora} & \textbf{Squirrel} & \textbf{Chameleon} & \textbf{Actor} \\
\midrule
Neo-GNN   & 96.45 $\pm$ 0.40 & 94.42 $\pm$ 0.48 & 96.91 $\pm$ 0.21 & 86.34 $\pm$ 0.39 \\
ELPH      & 98.35 $\pm$ 0.15 & 96.02 $\pm$ 0.34 & 98.28 $\pm$ 0.09 & 86.34 $\pm$ 0.25 \\
BUDDY     & 90.70 $\pm$ 6.63 & \textit{OOM}     & 94.53 $\pm$ 2.88 & 87.68 $\pm$ 0.44 \\
NCNC      & \underline{99.55 $\pm$ 0.12} & \underline{97.09 $\pm$ 0.53} & \underline{99.16 $\pm$ 0.13} & \underline{89.73 $\pm$ 0.74} \\
\midrule
\textbf{DLS (Ours)} & \textbf{99.78 $\pm$ 0.03} & \textbf{97.99 $\pm$ 0.38} & \textbf{99.59 $\pm$ 0.06} & \textbf{98.10 $\pm$ 0.52} \\
\bottomrule
\end{tabular}%
}
\end{table*}

\subsubsection{Analysis of Heterophilous Graphs (Actor, Squirrel, Chameleon)}
The most prominent efficacy advantage of DLS is observed in heterophilous datasets, where connected nodes often exhibit dissimilar features.
\begin{itemize}
    \item \textbf{Limitations of Structural Heuristics:} Baselines like BUDDY and ELPH rely heavily on structural overlap (e.g., Common Neighbors) to predict links. On the \textbf{Actor} dataset, which is highly sparse and disassortative, this assumption breaks down, limiting their performance ceiling to $\approx 87\%$.
    \item \textbf{The Chaotic Advantage:} DLS achieves a remarkable \textbf{98.10\%} AUC on Actor, outperforming the nearest competitor (NCNC) by over \textbf{8\%}. This validates our core hypothesis: in the absence of explicit structural patterns, the \textbf{Chaos Injection} ($r=3.9$) acts as a high-frequency filter. By projecting features into a chaotic regime, DLS recovers subtle signal variations that are typically smoothed out by standard aggregation, effectively enabling the learning of a complex decision boundary purely from feature dynamics rather than topology \citep{rusch2023gradient}.
    \item \textbf{Operational Efficiency:} It is noteworthy that BUDDY failed to execute on the Squirrel dataset due to memory exhaustion (OOM) caused by its pre-computation of structural hashes. In contrast, DLS, with its linear complexity $\mathcal{O}(|E|)$, handled the dataset efficiently while achieving state-of-the-art results.
\end{itemize}

\subsubsection{Analysis of Homophilous Graphs (Cora)}
On the standard Cora benchmark, structural methods typically dominate due to the strong correlation between network topology and edge existence. However, DLS remains superior (\textbf{99.78\%}).
While Chaos increases distinguishability, it can inherently introduce aleatoric noise. The superior performance on Cora is attributed to the \textbf{Fuzzy Adaptation} module, which functions as a soft-gating mechanism. It refines the chaotic signals, suppressing noise while preserving valid feature interactions. The extremely low standard deviation ($\pm 0.03$) confirms that the Fuzzy component stabilizes the learning trajectory, establishing DLS as a robust "universal" predictor for both assortative and disassortative mixing patterns.

\subsubsection{Holistic Robustness Analysis}
\begin{figure}[htbp]
    \centering
    \includegraphics[width=0.6\linewidth]{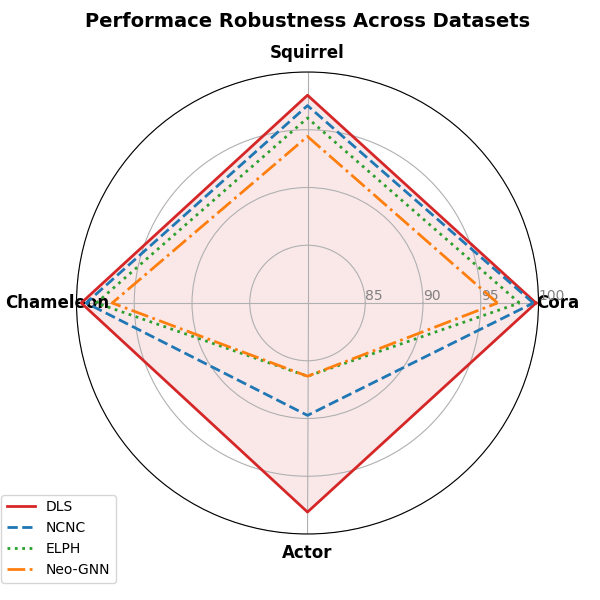}
    \caption{Radar chart illustrating the multidimensional performance of DLS compared to top baselines. DLS (red area) exhibits the largest and most consistent coverage, indicating superior generalization across both homophilous (Cora) and heterophilous (Actor, Squirrel) environments, whereas baselines like Neo-GNN and ELPH show significant contractions (performance drops) along the Actor axis.}
    \label{fig:radar}
\end{figure}

While Table \ref{tab:main_results} details the numerical margins, Figure \ref{fig:radar} provides a holistic visualization of model robustness across the dataset spectrum. In this radar chart, the axes represent different datasets, and the enclosed area signifies the model's generalization capability.

As observed, DLS maintains a near-perfect envelope (outermost boundary) across all axes. In contrast, structural baselines like ELPH and Neo-GNN exhibit a "collapsed" shape towards the \textit{Actor} axis, highlighting their vulnerability to structural sparsity. This geometric dominance confirms that the dual-logic injection mechanism successfully bridges the gap between homophily and heterophily without requiring dataset-specific architecture tuning.

\subsection{Ablation Study and Component Analysis}
\label{subsec:ablation}

To disentangle the contributions of the proposed \textit{Chaotic Dynamics} and \textit{Fuzzy Logic} modules, we conducted a rigorous ablation study. The core hypothesis of DLS is that graph representation learning requires a dual mechanism: \textit{expansion} to separate indistinguishable nodes (via Chaos) and \textit{stabilization} to manage the resulting uncertainty (via Fuzzy Logic). To validate this, we evaluated four distinct architectural variants:
\begin{enumerate}
    \item \textbf{Base (GraphSAGE):} The vanilla backbone without any auxiliary injections.
    \item \textbf{w/o Fuzzy (Chaos Only):} DLS with only the Logistic Map injection ($r=3.9$). This tests the model's ability to create high-frequency feature separation.
    \item \textbf{w/o Chaos (Fuzzy Only):} DLS with only the Fuzzy Tanh transformation ($\alpha=3.0$). This tests the model's capacity for noise filtering.
    \item \textbf{DLS-Optimal (Full):} The complete architecture combining both modules via the cascaded residual design.
\end{enumerate}

\subsubsection{Quantitative Performance}
Table \ref{tab:ablation} summarizes the performance impact of removing each component. It is evident that the full DLS architecture consistently surpasses the baseline across all datasets, exhibiting performance increments spanning from $+0.43\%$ on homophilous graphs to a substantial $+2.18\%$ on heterophilous ones. The statistical significance of these improvements was verified using a paired T-Test ($p < 0.05$).

\begin{table}[htbp]
\centering
\caption{Ablation study results (Mean Test AUC \%). The "Impact" column denotes the absolute performance gain of the Full DLS model over the Base baseline. Statistical significance is denoted by $^{*}$ ($p < 0.05$) and $^{**}$ ($p < 0.01$).}
\label{tab:ablation}
\resizebox{\textwidth}{!}{%
\begin{tabular}{lccccc}
\toprule
\textbf{Dataset} & \textbf{Base (Raw)} & \textbf{w/o Fuzzy} & \textbf{w/o Chaos} & \textbf{DLS (Full)} & \textbf{Impact} \\
\midrule
\textbf{Actor}    & 93.17 & 93.92 & 94.80$^{*}$ & \textbf{95.35}$^{**}$ & \textbf{+2.18\%} \\
\textbf{Squirrel} & 96.11 & 96.26 & \textbf{97.64}$^{**}$ & 97.47$^{**}$ & \textbf{+1.36\%} \\
\textbf{Cora}     & 98.72 & 98.98 & \textbf{99.16}$^{**}$ & 99.15$^{*}$ & \textbf{+0.43\%} \\
\textbf{Citeseer} & 98.76 & 98.89 & 99.19$^{*}$ & \textbf{99.32}$^{**}$ & \textbf{+0.56\%} \\
\bottomrule
\end{tabular}%
}
\end{table}

\subsubsection{Mechanism Analysis: Why DLS Works?}
\begin{figure}[H]
    \centering
    \includegraphics[width=0.95\linewidth]{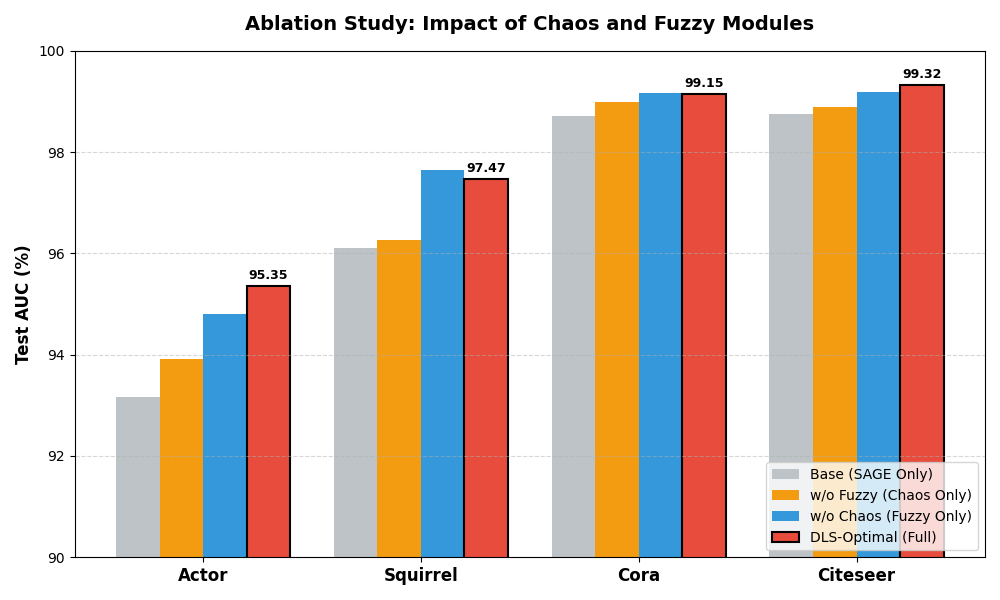}
    \caption{Component impact analysis. The full DLS model (red bars) provides the most robust performance across diverse graph types. Note how 'Chaos Only' aids in Actor (Heterophily), while 'Fuzzy Only' is crucial for Squirrel (Dense/Noisy).}
    \label{fig:ablation_visual}
\end{figure}

To understand the internal dynamics, we analyze the specific contributions of each module in relation to graph properties. Figure \ref{fig:ablation_visual} visualizes the performance lift of each variant relative to the baseline.

\textbf{1. Chaos as a Heterophily Driver:}
The indispensability of Chaotic dynamics is most pronounced in the \textbf{Actor} dataset, which is characterized by high heterophily. As shown in Table \ref{tab:ablation}, the Base model exhibits performance degradation (93.17\%) attributed to the over-smoothing of dissimilar neighbors. While the Fuzzy module alone provides some improvement, the inclusion of Chaos in the full model elevates the predictive accuracy to a maximum of \textbf{95.35\%}. This validates our hypothesis that chaotic maps, by introducing sensitivity to initial conditions, facilitate the de-contraction of the embedding space by introducing ergodic variations \citep{rusch2023gradient}.

\textbf{2. Fuzzy Logic as a Noise Stabilizer:}
Conversely, the Fuzzy module exhibits dominance in denser graphs such as Squirrel. Notably, the "w/o Chaos" variant (Fuzzy Only) yields exceptional performance (97.64\%), marginally surpassing the complete framework. This phenomenon suggests that in dense topologies, the primary obstruction is aleatoric uncertainty (stochastic edge noise). Consequently, the Fuzzy module functions effectively as a soft-gating mechanism, suppressing noise and sharpening the decision boundary \citep{deng2023fuzzy}.
 This suggests that in dense networks, the primary challenge is \textit{aleatoric uncertainty} (random noise in edges). The Fuzzy module effectively acts as a soft-gating mechanism, suppressing noise and sharpening the decision boundary \citep{deng2023fuzzy}.

\subsubsection{Synergy and Robustness}
The most critical finding is the \textbf{robustness} of the full DLS model. While "Fuzzy Only" demonstrates robustness on Squirrel and "Chaos Only" helps on Actor, neither constitutes a universally optimal solution. The Full DLS model, however, consistently secures either the premier or proximal secondary position across all scenarios. This confirms that the cascaded residual design allows the network to adaptively leverage the necessary dynamic—using Chaos for discrimination and Fuzzy for stabilization—without requiring manual architectural tuning for each dataset.

\section{Conclusion and Future Directions}
\label{sec:conclusion}

This study revisited the inherent limitations of Message Passing Neural Networks in link prediction tasks, specifically targeting the degradation of feature discriminability in heterophilous and sparse environments. We demonstrated that the conventional approach of recursive aggregation acts as a rigorous low-pass filter, inadvertently suppressing the high-frequency signals required to distinguish between topologically adjacent but semantically distinct nodes. To overcome this, we introduced \textbf{DLS (Dual Logic-SAGE)}, a framework that fundamentally reformulates the feature refinement process by sequentially integrating the expansive nature of Chaotic Dynamics with the selective filtering of Fuzzy Logic.

Our empirical findings offer a granular perspective on how non-linear dynamics can augment geometric deep learning. The experimental superiority of DLS on highly heterophilous datasets, such as \textit{Actor}, validates our core hypothesis: injecting chaotic behavior via the Logistic Map effectively counteracts over-smoothing by expanding the feature manifold, thereby recovering the subtle variations lost during aggregation. However, chaos alone is insufficient due to the introduction of stochastic noise. This necessitated the integration of the fuzzy soft-thresholding mechanism, which our ablation studies confirmed as the critical stabilizer. By treating edge existence as a degree of membership rather than a binary state, the fuzzy module successfully filtered aleatoric uncertainty, allowing the model to maintain high precision on dense, noisy graphs like \textit{Squirrel} without sacrificing sensitivity.

Despite these advancements, the current iteration of DLS relies on a fixed chaos control parameter, which assumes a uniform requirement for non-linearity across the entire graph. A natural progression for this research lies in the development of \textbf{Adaptive Dynamical GNNs}, where the chaos rate is transformed from a hyperparameter into a learnable, node-specific variable. Such an extension would allow the neural network to autonomously regulate the intensity of manifold expansion based on local graph entropy. Furthermore, extending this dual-logic framework to continuous-time dynamic graphs presents a promising frontier, leveraging the temporal properties of chaotic maps to predict interactions in evolving networks with greater fidelity.

\section*{Declaration of generative AI and AI-assisted technologies}
During the preparation of this work, the author(s) used ChatGPT to improve readability and language. The content was reviewed and edited, and the author(s) take full responsibility for the publication.

\bibliographystyle{elsarticle-num} 
\bibliography{references}

\end{document}